%% file: smooth_model_compression.tex
\documentclass{article}

\PassOptionsToPackage{numbers, compress}{natbib}

\usepackage[preprint]{neurips_2025}

\usepackage[utf8]{inputenc} % allow utf-8 input
\usepackage[T1]{fontenc}    % use 8-bit T1 fonts
\usepackage{xr-hyper}
\usepackage{hyperref}       % hyperlinks
\usepackage{url}            % simple URL typesetting
\usepackage{booktabs}       % professional-quality tables
\usepackage{amsfonts}       % blackboard math symbols
\usepackage{nicefrac}       % compact symbols for 1/2, etc.
\usepackage{microtype}      % microtypography
\usepackage{xcolor}         % colors
\usepackage{wrapfig}

\input{preamble}

\title{Smooth Model Compression without Fine-Tuning}

\author{%
  Christina Runkel\thanks{Equal contribution.} \\
  University of Cambridge \\
  \texttt{cr661@cam.ac.uk} \\
  \And
  Natacha Kuete Meli* \\
  University of Siegen \\
  \AND
  Jovita Lukasik \\
  University of Siegen \\
  \And
  Ander Biguri \\
  University of Cambridge \\
  \And
     Carola-Bibiane Sch{\"o}nlieb \\
  University of Cambridge \\
  \AND
  Michael Moeller \\
  University of Siegen \\
}

\begin{document}

\maketitle

\input{sec/0_abstract}
\input{sec/1_introduction}
\input{sec/2_related_work}
\input{sec/3_method}

\input{sec/4_experiments}

\input{sec/5_conclusion}
{
     \small
     \bibliographystyle{ieeenat_fullname}
     \bibliography{smooth_model_compression}
 }

\input{sec/6_appendix}

%%%%%%%%%%%%%%%%%%%%%%%%%%%%%%%%%%%%%%%%%%%%%%%%%%%%%%%%%%%%

% \input{sec/7_checklist}

\end{document}

%% file: preamble.tex
\usepackage{graphicx}
\usepackage{tikz}
\usepackage{quantikz}
\usetikzlibrary{positioning,patterns,3d,angles,quotes,shapes.arrows,fadings}
\usepackage{pgfplots}
\pgfplotsset{compat=1.18}
\usepackage{upgreek}
\usepackage{ifthen}
\usepackage{calc}
\usepackage{amsmath,amssymb,amsthm,times, bm}
\usepackage{hyperref}
\usepackage{cleveref}
\usepackage{bbm}
\usepackage{graphicx}
\usepackage{multicol} 
\usepackage{multirow}
\usepackage{caption}% for figure descriptions
\usepackage{subcaption}% for subfigure descriptions
\usepackage{braket}
\usepackage{pifont}
\usepackage{fontawesome}
\usepackage{xifthen} % For conditional statements
\usepackage{algorithm}
\usepackage{algpseudocode}

\usepackage{wrapfig}
\usepackage{comment}
\usepackage[framemethod=tikz]{mdframed}

\newcommand{\R}{\mathbb{R}}

%% file: sec/0_abstract.tex
\begin{abstract}
Compressing and pruning large machine learning models has become a critical step towards their deployment in real-world applications. Standard pruning and compression techniques are typically designed without taking the structure of the network's weights into account, limiting their effectiveness. 
We explore the impact of smooth regularization on neural network training and model compression. 
By applying nuclear norm, first- and second-order derivative penalties of the weights during training, we encourage structured smoothness while preserving predictive performance on par with non-smooth models. 
We find that standard pruning methods often perform better when applied to these smooth models. 
Building on this observation, we apply a Singular-Value-Decomposition-based compression method that exploits the underlying smooth structure and approximates the model's weight tensors by smaller low-rank tensors. 
Our approach enables state-of-the-art compression without any fine-tuning -- reaching up to 91\% accuracy on a smooth ResNet-18 on CIFAR-10 with 70\% fewer parameters.
\end{abstract}

%% file: sec/1_introduction.tex
\section{Introduction}
\label{sec:introduction}
In recent years, deep neural networks have shown to be highly successful for real-world applications. A large part of these successes is the increasing number of parameters of those networks, making them powerful tools for many tasks in computer vision and natural language processing. However, with an increasing number of parameters, training and inference on those neural networks becomes more and more computationally expensive. 
Model compression techniques mitigate this issue by removing or combining redundant parameters (model pruning or compression), reducing the precision of the parameters (quantization) or training student networks to imitate the behavior of the larger teacher network (knowledge distillation) without significantly degrading its performance \cite{cheng2017survey}. 

Previous works \cite{solodskikh2023integral, feinman2019learning} have shown that smoothness of the network's weights is improving pruning results, leading to a greater reduction of the size of the network. While Feinman and Lake \cite{feinman2019learning} focus on smoothness in spatial dimension, motivated by the smoothness of receptive fields in the primary visual cortex, Integral Neural Networks (INNs) \cite{solodskikh2023integral} enforce smoothness in channel dimension. In INNs the weights of a neural network are represented as continuous functions. The forward pass is computed via numerical integration quadratures, transforming the continuous representations into discrete formulations. Computing these numerical integral quadratures, however, is computationally expensive. We thus propose \textit{smooth weight learning}. Smooth weight learning is a computationally cheap alternative to other smoothing techniques like INN \cite{solodskikh2023integral}. It enforces smoothness in output channel dimension by adding a regularization term to the loss function during training. This allows the development of pruning methods exploiting the smooth structure of the weights. We therefore introduce \textit{singular value decomposition compression} - a fast pruning method that approximates the network's weight tensors by low-rank matrices. 

Our \textbf{contributions} are:
\begin{itemize}
    \item We introduce \textit{smooth weight learning}. It enforces smoothness of the network's weights by adding a regularization term to the loss function during training. For a suitable smoothing factor, our method improves the accuracy on downstream tasks like CIFAR10 classification compared to a network trained without smooth weight learning while also enabling the use of tailored pruning techniques. 
    \item Exploiting the smooth structure of the weights in output dimension, we propose to use \textit{Singular value decomposition compression} (SVD compression), a fast pruning method that reduces the network's number of parameters by replacing its weight matrices by low-rank approximations. 
    \item We show the effectiveness of smooth weight learning in combination with SVD Compression on the example of implicit neural representation learning and CIFAR10 classification.
\end{itemize}

%% file: sec/2_related_work.tex
\section{Related work}
\label{sec:related_work}
\paragraph{Model compression.}
Model compression techniques cluster into quantization, knowledge distillation and pruning methods \cite{cheng2017survey}. \textit{Quantization} methods like \cite{krishnamoorthi2018quantizing, hubara2018quantized, dong2019hawq, zhou2016dorefa} reduce the precision of the network's parameters to reduce memory usage and speed up computation time. \textit{Knowledge distillation} approaches \cite{hinton2015distilling, gou2021knowledge, romero2014fitnets, furlanello2018born, zhang2019your, touvron2021training} train a smaller student network with the aim of imitating the behavior of the deeper teacher network.

\textit{Model pruning} methods can be classified into unstructured- and structured pruning methods. While unstructured pruning methods like \cite{yang2024neural, wu2023model, mitsuno2021filter, han2015learning, lecun1989optimal, hassibi1993optimal, entezari2021role, frankle2018lottery, frantar2022optimal, he2018multi} prune the model's weights independent of any underlying structure, structured methods prune whole structures like channels or layers. This makes them more efficient in terms of computational and memory resources as they reduce the size of the weight tensor, freeing up memory and saving floating point operations. Due to the nature of unstructured pruning methods, decreasing the size of the weight tensors is usually impossible, leaving them with setting single parameters to zero. While this can free up memory and enable the use of specific algorithms for sparse matrices, setting weight parameters to zero usually does not necessarily reduce computation time. Unstructured pruning methods, however, commonly interfere less with model performance than structured pruning approaches as replacing cohesive structures tends to be more challenging than making up for isolated parameters. This especially holds for earlier layers as the error propagates through the rest of the network. More traditional structured pruning approaches include \cite{wen2016learning, li2016pruning} and \cite{hu2016network}. Structured pruning methods focus on identifying structures of convolutional layers and replace whole channels (see e.g., \cite{ye2018rethinking, peng2019collaborative}), make use of optimal transport theory \cite{theus2024towards} or identify global structures and correlations for pruning \cite{you2019gate, nonnenmacher2021sosp}. With the recent rise in popularity of large language models (LLMs), more recent pruning techniques like SliceGPT \cite{ashkboos2024slicegpt}, ShortGPT \cite{men2024shortgpt} and SparseGPT \cite{frantar2023sparsegpt} tackle the problem of reducing the size of LLMs. Further LLM pruning methods include those of Kwon et al. \cite{kwon2022fast}, Sun et al. \cite{sun2023simple} and An et al. \cite{an2024fluctuation}. 

While traditional pruning methods (see e.g., \cite{liu2024lightweight, he2017channel, luo2017thinet}) focused on pruning during training of the network, pruning without fine-tuning, i.e., after training, has become more popular in recent years \cite{kwon2022fast, frantar2023sparsegpt, sun2023simple, an2024fluctuation}. Pruning without fine-tuning saves time as it avoids the compute and time costs of retraining or fine-tuning. It also allows for a clear distinction between training and pruning method, making it more versatile. Pruning without fine-tuning, however, is known to be more challenging especially for high compression as the decline in accuracy is usually larger than for pruning methods with fine-tuning or retraining. The current state of the art structured pruning method that does not require any fine-tuning on data is model folding \cite{wangforget}. Model folding merges structurally similar neurons via clustering of filters. The data statistics of the compressed model are then corrected via an additional data- and fine-tuning free repair step, addressing the issue of variance collapse of data statistics in compressed models.

\paragraph{Smoothness through regularization.}
Enforcing smoothness through regularization is a well-known concept in the machine learning community. Methods like double backpropagation \cite{drucker1992improving} regularize the output of the network. Various other regularization techniques like weight decay \cite{loshchilov2017decoupled} (regularizing with an $\ell_2$ term) or variants thereof (e.g., using the $\ell_1$ norm instead) are commonly used to improve training results. A specific focus is put on spectral normalization techniques \cite{miyato2018spectral, runkel2021depthwise, sedghi2018singular, yoshida2017spectral} which control the Lipschitz constant of the layer. Enforcing a small spectral norm of the weights improves generalization of the network, stabilizes the training procedure, yields invertibility of the layers \cite{behrmann2019invertible} and enables continual learning \cite{kirkpatrick2017overcoming}. Other methods to enforce global smoothness through Lipschitz constraints include \cite{gouk2021regularisation, tsuzuku2018lipschitz, cisse2017parseval} and \cite{bartlett2018representing}. 
In addition to regularizing the weights of the network, regularizing the norm of the gradient with respect to the inputs has shown to increase training stability \cite{sokolic2017robust, gulrajani2017improved, fedus2018many, arbel2018gradient, kodali2017convergence}.
%Yet, a regularization via the spectral norm, i.e., the largest singular value, typically leads to full rank matrices, because the accumulating the expressiveness in a few (albeit large) singular values is penalized. For applications like model compression a low-rank weight matrix is often desired for exploiting facotrizations. 
Interestingly, the idea to enforce smoothness along particular dimensions of network parameters (e.g., of the 4-dimensional parameter tensor of convolutions with two spatial, one input-channel, and one output-channel dimension) does not seem to be explored, despite encouraging recent results of such smoothness on compression in the context of integral neural networks \cite{solodskikh2023integral}.

\paragraph{Smoothness via integral neural networks.} Integral neural networks (INNs) \cite{solodskikh2023integral} have proposed to view the weights of a layer as (continuous) functions, and the action of each linear layer as a suitable integral. They propose to discretize the resulting integrals via a different Monte-Carlo sampling in each forward pass, inherently avoiding the permutation invariance of classical linear layers and implicitly leading to smoothness along all dimensions of the weights that are viewed continuously. While the weight-smoothness is demonstrated to have favorable effects on dedicated compression methods, the  Monte-Carlo sampling requires computing a different interpolation from the underlying discretization for each forward pass.  In this work, we demonstrate that a similar effect can be achieved \textit{without} the interpolation introduced by the continuous interpretation using a suitable smoothness penalty on the discrete weight tensors directly. 

%% file: sec/3_method.tex
\section{Method}
\label{sec:method}
In this section, we introduce \textit{smooth weight learning} and \textit{singular value decomposition (SVD) compression}. Smooth weight learning enforces smoothness of the weights of the model in the output dimension of each layer by adding a regularization term during training. SVD compression exploits the smooth structure of the model's weight tensors to replace them by smaller low-rank ones. 

\begin{figure}[t]
    \centering
    \includegraphics[trim=0pt 0pt 0pt 19pt, clip, width=0.8\linewidth]{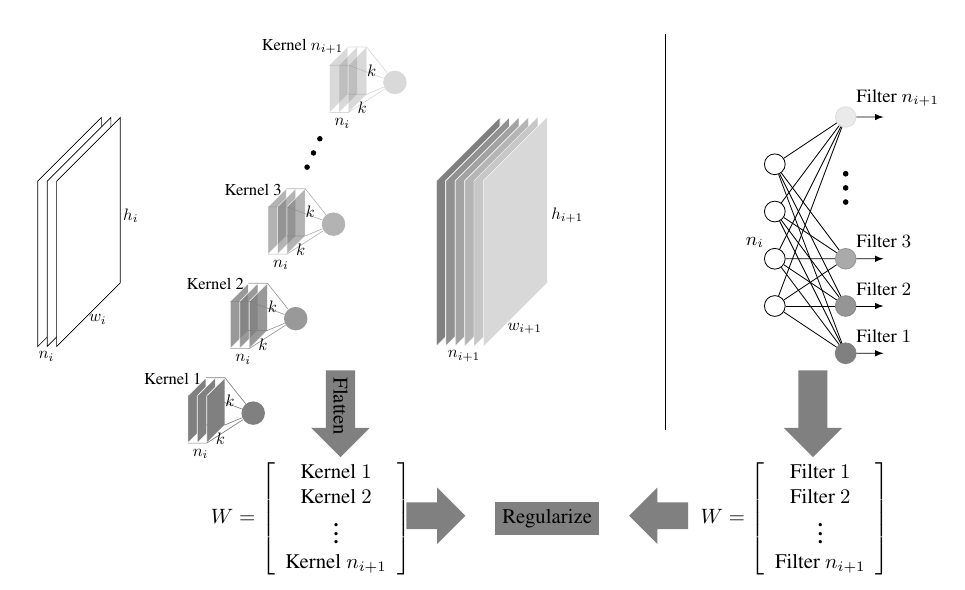}
    \caption{Illustration of our \textit{smooth weight learning} in output channel dimension for \textbf{(left)} a convolutional layer, and \textbf{(right)} a linear layer.
    For the convolutional layer, we flatten the kernels in the rows of a weight tensor $W$, which become similar to the weight tensor of the linear layer.
    Regularizing the output channel dimension then yields penalizing a function of the rows of $W$.
    }
    \label{fig:smooth_visualization}
\end{figure}

\subsection{Smooth weight learning}
\label{sec:soomth_weight_learning}
We introduce smooth weight learning, a weight smoothing technique regularizing the weights of a neural network during training. Figure \ref{fig:smooth_visualization} gives an overview of smooth weight learning for both convolutional and linear layers. For convolutional layers, the weight tensors are flattened such that their structure resembles that of a linear layer. With the aim of achieving smoothness in the output dimension of the layers of a network, we then add a regularization term $\mathcal{R}$. Tailored towards the SVD compression, we try regularizing with the nuclear norm. Computing the nuclear norm requires computing the singular values of $W$ and therefore is computationally expensive. We additionally try adding first and second order derivative penalties to the loss function during training, enforcing local smoothness. For an arbitrary loss function $l$, an input $x$, a smoothing factor $\lambda$ and a neural network $\Phi$ parametrized by weights $W$, the updated loss function $\mathcal{L}$ thus reads:
\begin{equation}
    \mathcal{L}(x) := l\big(\Phi(W; x)\big) + \lambda \mathcal{R}(W).
\end{equation}
\paragraph{Nuclear norm regularization term.}
A nuclear norm is the convex envelope of the rank function on the unit ball of matrices, i.e., it can be used as measure of complexity of a matrix. Regularizing with a nuclear norm term thus penalizes matrices that have a high rank while favoring those that are close to low-rank. With regard to pruning methods, low-rank matrices that concentrate most of its information in a few dimensions are favorable to retain good results on downstream tasks while decreasing the number of model parameters and thus memory and computation costs. 
For singular values $\sigma$, we define the nuclear norm regularization term as:
\begin{equation}
    \mathcal{R}_{\text{nuc}} (W) := \frac{1}{N \cdot m} \sum_{i=1}^{m} \sigma_i(W).
\end{equation}

\paragraph{First order derivative regularization term.}
To enforce smoothness of output channels of convolutional layers and in the output dimension of linear layers, we penalize large differences in neighboring output channels and dimensions, respectively. We define a first order derivative regularization term
\begin{equation}
    \mathcal{R}_1(W) := \frac{1}{N \cdot (n_o-1)} \sum_{i=1}^N \sum_{j=1}^{n_o -1} \big\|W_{i, j} - W_{i, (j+1)}\big\|_1
\end{equation}
for weights $W$, $N$ and $n_o$ being the number of layers of the network and dimensionality of the output dimension, respectively. During training, we flatten the weight matrices of convolutional layers.

\paragraph{Second order derivative regularization term.}
Regularizing with a first order derivative term not only enforces smoothness but also penalizes linear dependencies between output dimensions. With regard to model compression and pruning, however, linear dependencies are favorable as they enable e.g., approximating a channel by a linear combination of its neighboring channels. We thus additionally introduce a second derivative regularization term that allows for linear dependencies between layers while enforcing smoothness in output dimension direction, i.e.,
\begin{equation}
    \mathcal{R}_2(W) := \frac{1}{N \cdot (n_o -2)} \sum_{i=1}^N \sum_{j=1}^{n_o -2} \big\|W_{i, j} - 2W_{i, (j+1)} + W_{i, (j+2)}\big\|_1.
\end{equation}

Smooth weight learning with first- and second order derivative penalty terms has the advantage of being computationally cheap as it only requires computing $N \cdot (n_o -1)$ and $N \cdot (n_o -2)$ additional $\ell_1$ norms for the first- and second order derivative regularization terms, respectively.\footnote{It is worth noting that while we only exploit smoothness in output dimension as it is most relevant for pruning and SVD compression, smooth weight learning can be extended to any dimension of the weight tensor.}

\subsection{Singular value decomposition compression}
\label{sec:svd_compression}

Given a matrix $W \in \mathbb{C}^{m \times n} $ and a desired rank $r\in [0, \min\{m, n\}]$, the low-rank approximation minimizes the Frobenius norm of the reconstruction error:
\begin{equation}
    \| W - \widetilde{W} \|_F = \min_{\text{rank}(\hat{W}) \leq r} \| W - \hat{W} \|_F.
\end{equation}
It is known \cite{kishore2017literature} that the rank-$r$ matrix $\widetilde{W}^\star$ obtained from the truncated singular value decomposition (SVD) of $W$ provides the best rank-$ r $ approximation of $ W $ in terms of reconstruction fidelity.

The SVD of this matrix $W$ can be expressed as 
\begin{equation}
    W = U \Sigma V^\top,
\end{equation}  
where $ V \in \mathbb{C}^{n \times n} $ is an orthogonal matrix whose columns are the \emph{right singular vectors} of $ W $, forming an orthonormal basis of the input feature space;  
$ \Sigma \in \mathbb{C}^{m \times n} $ is a diagonal matrix whose non-negative entries are the \emph{singular values} of $ W $, representing the strength of each corresponding mode;  
and $ U \in \mathbb{C}^{m \times m} $ is an orthogonal matrix whose columns are the \emph{left singular vectors} of $ W $, forming an orthonormal basis of the output feature space.

Since the singular values in $ \Sigma $ are typically ordered in descending magnitude, it is often observed that only a small number of them carry most of the energy or information content of the original matrix, and this specially if $W$ is structured. 
Let then $ r < \min(m, n) $ denote a target rank. 
A rank-$ r $ approximation of $ W $ can then be constructed by truncating the SVD to its top $ r $ components, yielding
\begin{equation}
\label{eq:svd}
    W \approx \widetilde{W} = U_r \Sigma_r V_r^\top,
\end{equation}
where $ U_r \in \mathbb{C}^{m \times r} $, $ \Sigma_r \in \mathbb{C}^{r \times r} $, and $ V_r \in \mathbb{C}^{n \times r} $ consist of the first $ r $ columns of $ U $, the top $ r $ singular values, and the first $ r $ columns of $ V $, respectively.
Following \cite{acharya2019online,yu2017compressing}, we aim to use this low-rank approximation to reduce the size of the weights matrices on the neural network.

\paragraph{Linear layer.}
Let $n_i$ denote the number of features of the input to the $i$th linear layer of the model.
The weight tensor of this layer is a matrix $W \in \R^{n_{i+1}, n_i}$.
The SVD of this matrix $W$ using Eq. \eqref{eq:svd} can be expressed as $ \widetilde{W} = U_r \Sigma_r V_r^\top $,
where $ U_r \in \mathbb{C}^{n_{i+1} \times r} $, $ \Sigma_r \in \mathbb{C}^{r \times r} $, and $ V_r \in \mathbb{C}^{n_i \times r} $.

We use this approximation to replace the original tensor $W$ with two smaller tensors $W_1 = \Sigma_r V_r^\top \in \mathbb{C}^{n_{i+1} \times r}$ and $W_2 = U_r \in \mathbb{C}^{r \times n_i}$, yielding two consecutive linear layers indexed by $i_1$ and $i_2$.
This yields the mapping $x \mapsto W_2(W_1 x)$, where the input $ x \in \mathbb{C}^{n_i} $ is first projected into a low-dimensional subspace, scaled, and then mapped back into the output space.
For completeness, we copy the bias vector of the original linear layer $i$ in that of the added linear layer $i_2$ and set the bias vector of $i_1$ to $0$.

The added linear layers $i_1$ and $i_2$ count a total of $ r(n_i + n_{i+1})$ parameters while the original layer $i$ counts $n_i \cdot n_{i+1}$ parameters, up to biases.
Given a target sparsity $s \in [0, 1]$, we determine the rank $r$ that matches the target sparsity $s$ by solving the following equation for $r$ and rounding the result:
\begin{equation}
    s =  1 - \frac{r(n_i + n_{i+1}) + n_{i+1}}{n_i \cdot n_{i+1} + n_{i+1} }.
\end{equation}

\paragraph{Convolutional layer.}
Let $ n_i $ denote the number of input channels of spatial dimension $ (h_i, w_i) $ to the $ i $th convolutional layer of the model. 
The weight tensor of this layer is a 4D tensor $ W \in \mathbb{C}^{n_{i+1} \times n_i \times h_i \times w_i} $, where $ n_{i+1} $ is the number of output channels.
Interestingly, the convolution operation can be equivalently expressed as a matrix–vector multiplication. 
Specifically, by unfolding the input feature map, the convolution becomes a matrix multiplication between a reshaped weight matrix and the flattened input patches. 
This reshaping transforms the weight tensor $ W $ into a matrix $ \tilde{W} \in \mathbb{C}^{n_{i+1} \times (n_i h_i w_i)} $, where each row corresponds to a kernel, and each column corresponds to a position in the receptive field.

A low-rank approximation of this matrix $\tilde{W}$ using in Eq. \eqref{eq:svd} yields $\widetilde{W} \approx U_r \Sigma_r V_r^\top$, where $ U_r \in \mathbb{C}^{n_{i+1} \times r} $, $ \Sigma_r \in \mathbb{C}^{r \times r} $, and $ V_r \in \mathbb{C}^{n_i h_i w_i \times r} $.
This allows to replace the original weight tensor $\tilde{W}$ with two low-rank tensors $\widetilde{W}_1 = \Sigma_r V_r^\top \in \mathbb{C}^{r \times n_i h_i w_i}$ and $\widetilde{W}_2 =  U \in \mathbb{C}^{n_{i+1} \times r} $, the first projecting and scaling the input into a low-dimensional subspace and the second reconstructing it to the output space.
We now unfold the new tensors to $W_1 \in \R^{r \times n_i \times h_i \times w_i}$ and $W_2 \in R^{n_{i+1}, r, 1, 1}$, yielding two consecutive linear convolution operations layers by $i_1$ and $i_2$.
We copy the bias vector of the original convolutional layer layer $i$ in that of the added linear $i_2$ and set the bias vector of $i_1$ to zero.

The added layers count a total number of $ r(n_i h_i w_i + n_{i+1}) $ parameters while the original layer counts $ n_{i+1} \cdot n_i \cdot h_i \cdot w_i $, up to biases.
Given a target sparsity $s \in [0, 1]$, we determine the rank $r$ that matches the target sparsity $s$ by solving the following equation for $r$ and rounding the result:
\begin{equation}
    s = 1 - \frac{r(n_i h_i w_i + n_{i+1}) + n_{i+1}}{n_{i+1} \cdot n_i \cdot h_i \cdot w_i + n_{i+1}}. 
\end{equation}

%% file: sec/4_experiments.tex
\section{Numerical experiments}
\label{sec:experiments}
To demonstrate the effectiveness of our methods, we perform experiments for implicit neural representation learning and image classification. We compare our smooth models to integral neural networks \cite{solodskikh2023integral}, and our pruning method, i.e., our combination of smooth models and SVD-compression, to traditional structured- and unstructured pruning methods as well as model folding \cite{wangforget}.
\subsection{Implicit neural representations}
\begin{figure}[htb]
    \centering
        \begin{tabular}{p{4.3cm}p{4.3cm}p{4.3cm}}
        \includegraphics[width=1\linewidth]{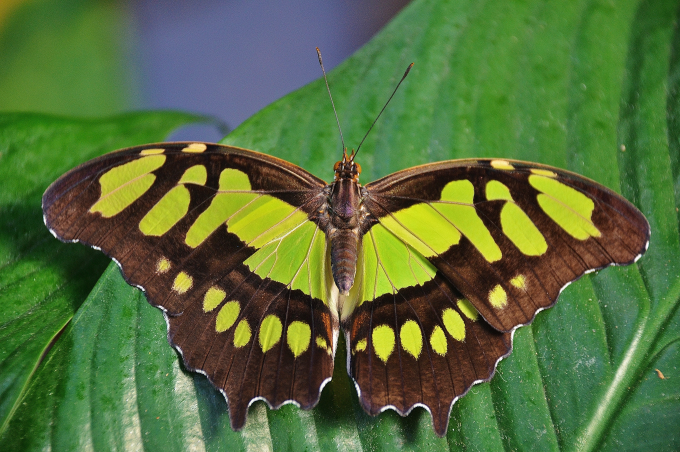} &  \includegraphics[width=1.0\linewidth]{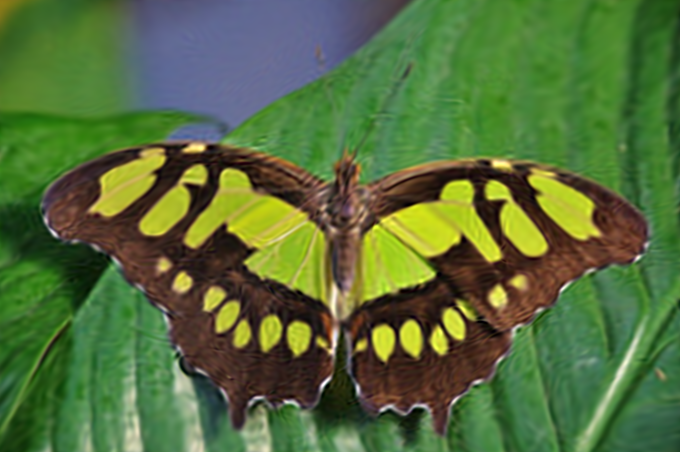}& \includegraphics[width=1.0\linewidth]{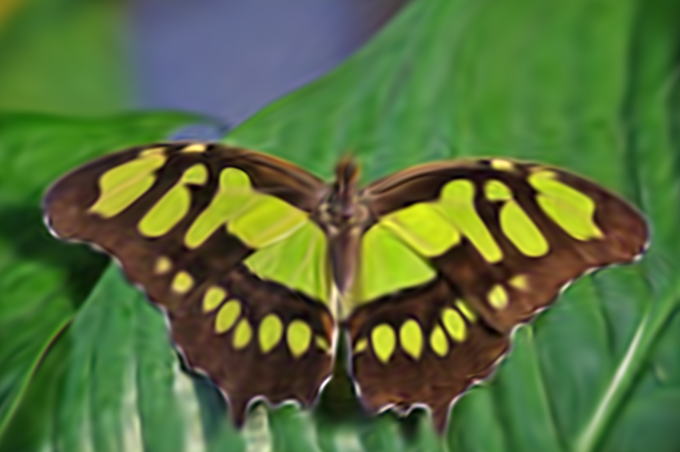} \\
            Original Image, 922,080 pixel values & $\mathcal{R}_{\text{nuc}}$, $\lambda = 0.001$, no SVD compression, 133,958 parameters, PSNR=26.8 & $\mathcal{R}_{\text{nuc}}$, $\lambda = 0.01$, and $49\%$ SVD compression, 68,074 parameters, PSNR=25.9
        \end{tabular}
    \caption{Original image (left) and single image super resolution (SISR) reconstructions from a 4x downsampled version using a nuclear norm regularization (middle), and an additional SVD-based model compression without finetuning (right). }
    \label{fig:qualitative_sisr}
\end{figure}

Implicit neural representations are implicitly defined continuous representations of an input signal or image. Using multilayer perceptrons (MLP) to approximate the underlying continuous function, they have shown to be powerful tools especially when dealing with irregularly sampled data \cite{eslami2018neural, sitzmann2020implicit, mildenhall2021nerf, saragadam2023wire}.
\begin{wrapfigure}{r}{0.45\textwidth}
  \begin{center}
\includegraphics[width=0.45\textwidth]{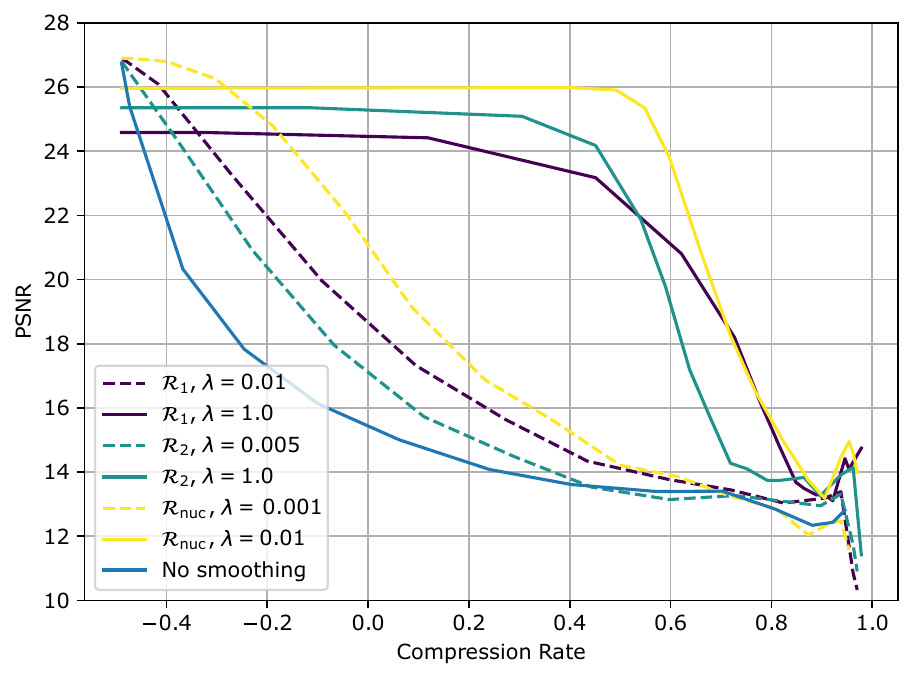}
  \end{center}
  \vspace{-0.5cm}
 %\caption{Single image super resolution on the "butterfly" image with a subsequent SVD-based compression.}
    \label{fig:inr_results}
\end{wrapfigure}
 As a proof of concept, we run the single-image super-resolution experiment of WIRE \cite{saragadam2023wire} using the code provided by the authors. It reconstructs an RGB image (from a 4x subsampled version) using two hidden layers with $181 \times 181$ complex-valued parameters (corresponding to the same expressiveness as $256\times 256$ real parameters) and complex wavelet activations. In addition to the data term, we regularize all linear layers with $  \mathcal{R}_1 $, $  \mathcal{R}_2 $, and $  \mathcal{R}_{\text{nuc}} $ with penalty parameters $\lambda \in \{0, 0.001, 0.005, 0.01, 0.05, 0.1, 0.5, 1, 5\}$. We then apply a joint SVD compression of the stacked weight matrices of the two hidden layers. 
  We illustrate the PSNR of the represented image for different compression rates and two choices of $\lambda$ per regularizer in the figure on the right. Please note that the factorization of weights into the product of two matrices can lead to a parameter increase if the rank remains large. This explains the visualization of negative compression rates (although the resulting network can of course be stored at no extra costs). While small regularization parameters lead to reconstructions on par (or slightly superior to) unregularized reconstructions, a large nuclear norm regularizer allows compressing the resulting implicit representation by almost $50\%$ without significant loss in image quality. This is particularly impressive considering that the (uncompressed) implicit representation itself only has about $7$ times fewer parameters than the images has pixel values. Figure \ref{fig:qualitative_sisr} shows qualitative results of the resulting images. 

\subsection{Image classification}
\label{sec:results_image_classification}
To show the effectiveness of our method on a second downstream task, we train a ResNet18 \cite{he2016deep, he2019bag} model for CIFAR10 \cite{krizhevsky2009learning} classification. The hyperparameters are chosen based on \cite{geiping2022stochastic}. The model is trained with a batch size of 128 for 300 epochs using SGD with a learning rate of $0.01$, a weight decay of $10^{-4}$ and Nesterov momentum of $0.9$. The learning rate is warmed up from $0.0$ to $0.1$ over the first five epochs. After this, we use cosine annealing to reduce it back to $0$ over the remaining 295 epochs.

Table \ref{tab:smooth_weight_learning_CIFAR10} provides an overview of the results for smooth weight learning with smoothing factors $\lambda \in \{0.01, 0.05, 0.1, 0.2, 0.3, 0.4, 0.5, 1.0, 2.0, 2.5, 5.0, 7.5, 15.0 \}$ using first- and second order regularization terms $\mathcal{R}_1$ and $\mathcal{R}_2$, respectively. We compute the test accuracy on the official CIFAR10.

\begin{table}[t]
    \centering
    \resizebox{\textwidth}{!}{
    \begin{tabular}{c |c | c |c|c|c|c|c|c|c|c|c|c|c|c|c }\hline
          \multicolumn{2}{c|}{$\lambda$} & \textbf{0.0} & \textbf{0.01} & \textbf{0.05} & \textbf{0.1} & \textbf{0.2} & \textbf{0.3} & \textbf{0.4} & \textbf{0.5} & \textbf{1.0} & \textbf{2.0} & \textbf{2.5} & \textbf{5.0} & \textbf{7.5} & \textbf{15.0} \\ \hline
          \multirow{2}{*}{\rotatebox{90}{\textbf{Acc}}} & $\mathcal{R}_1$ & 94.14 & 94.04 & \underline{\textbf{94.51}} & 94.03 & 94.08 & 94.38 & 94.06 & 93.97 & 94.03 & 93. 63 & 93.40 & 93.23 & 92.71 & 91.89 \\
          & $\mathcal{R}_2$ & 94.14 & 94.43 & 94.22 & \underline{\textit{94.40}} & 94.10 & 94.21 & 94.08 & 94.01 & 93.78 & 93.61 & 93.55 & 93.24 & 93.00 & 92.28\\
    \end{tabular}}
    \caption{Accuracy on CIFAR10 test set for a ResNet18 model for first- and second order derivative regularization terms ($\mathcal{R}_1$ and $\mathcal{R}_2$, respectively) for smoothing with an $\ell_1$ norm and different smoothing factors $\lambda$. Training the network without smooth weight learning leads to an accuracy of $94.14\%$. The same network trained with a smoothing factor $\lambda=0.05$ and a first order derivative regularization term improves the results by nearly $0.4\%$ to $94.51\%$. For a suitable smoothing factor, smooth weight learning improves the accuracy for both the first- and second order derivative penalty terms.}
    \label{tab:smooth_weight_learning_CIFAR10}
\end{table}
Compared to the same network trained without smoothing, smooth weight learning improves the accuracy for a suitable smoothing factor by up to $0.4\%$ for both a first- and second order derivative regularization term. For large smoothing factors, computing the accuracy before pruning leads to a drop in accuracy by $2.25$ and $1.86$ percent for a first- and second order regularization term. Regularizing with the nuclear norm, i.e., the $\mathcal{R}_{\text{nuc}}$ regularization term, slightly improves the accuracy for small smoothing factors, but drops off quickly to only $74\%$ for large smoothing factors. As smoothness of the weights of the network has shown to be beneficial for pruning, in the following we investigate the impacts when pruning with SVD compression. Due to  the large drop in accuracy using  $\mathcal{R}_{\text{nuc}}$, in the following, we consider $\mathcal{R}_{1}$ and  $\mathcal{R}_{2}$ only, and provide more results on regularizing with the nuclear norm in the appendix.

Figure \ref{fig:svd_inspection} highlights the cumulative percentage of singular values per layer of the trained ResNet18 model for different smoothing factors and their singular value index for training with an $\mathcal{R}_1$ regularizer (top row) and $\mathcal{R}_2$ regularizer (bottom row). The singular values are ordered with descending magnitude. As we are interested in approximating the original weight matrix $W$ by $r$ (non-zero) singular values, we aim to train a network where the cumulative percentage of the singular values of each layer is as high as possible while keeping the singular value index as low as possible. 
For a model trained without smooth weight learning, reducing the rank of the matrix to 100 results in less than $45\%$ of the singular values being preserved. Reducing the rank of the same network architecture trained with
smooth weight learning and a smoothing factor $\lambda = 15.0$ to 100, more than $60\%$ and $70\%$ of the singular values are preserved for $\mathcal{R}_1$ and $\mathcal{R}_2$ regularization terms, respectively. Preserving $15$ and $25\%$ more of the singular values indicates better performance on the downstream task for higher levels of sparsity. With increasing smoothing factor, the cumulative percentage of preserved singular values when cropping at the same singular value index increases - confirming the trend.
\begin{figure}
    \centering
    \includegraphics[width=1.\linewidth]{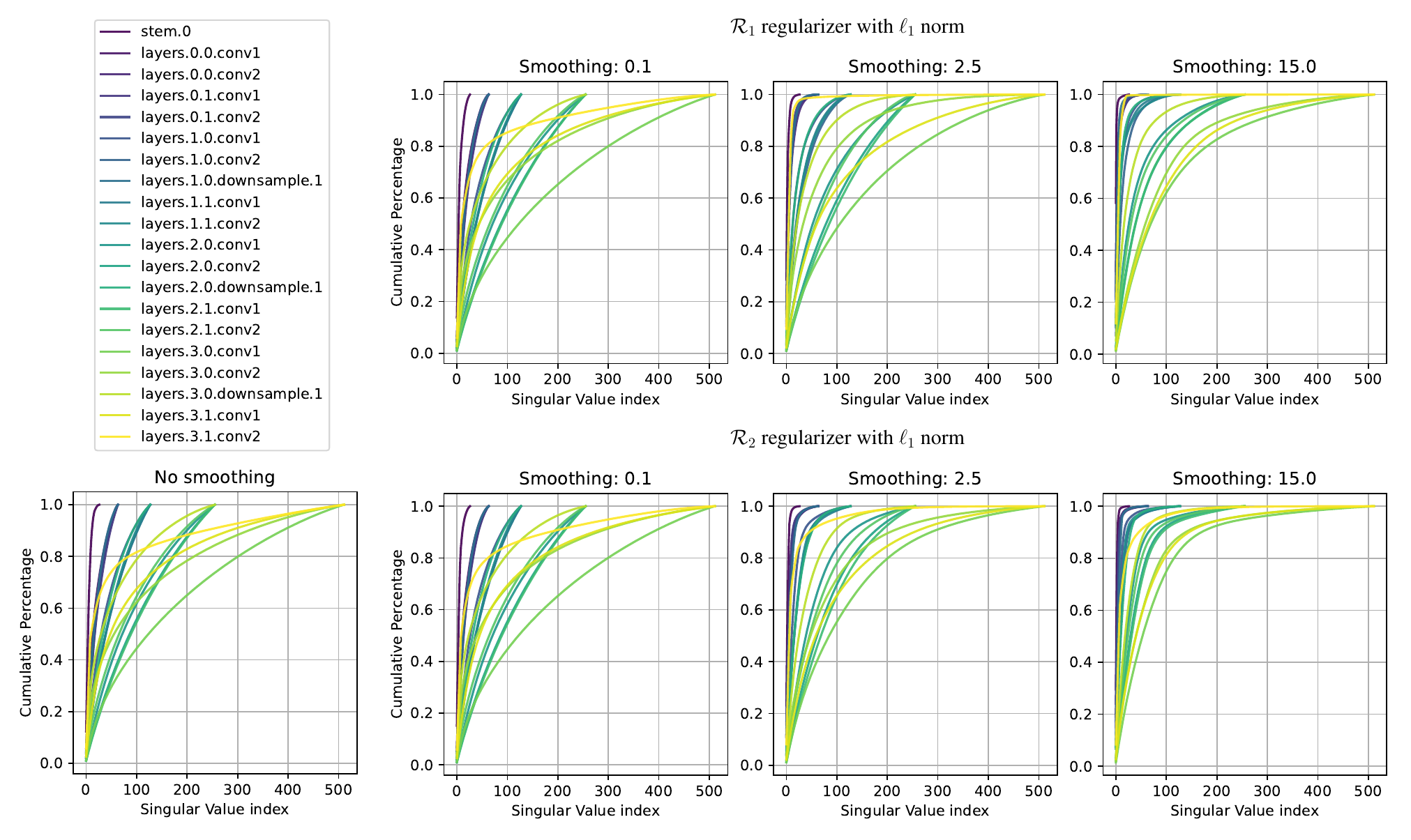}
    \caption{Inspecting the singular values of the smooth Resnet18 layers for different smoothing factors.
    For reference, the singular values for a non-smooth Resnet18 are shown (bottom left).
    We see that both for the $\mathcal{R}_1$ and $\mathcal{R}_2$ regularizer, increasing the smoothing factor results in clustering the singular values of the weight tensor, which is favorable to the SVD compression.
    }
    \label{fig:svd_inspection}
\end{figure}

\begin{figure}
    \centering
    \includegraphics[width=1.\linewidth]{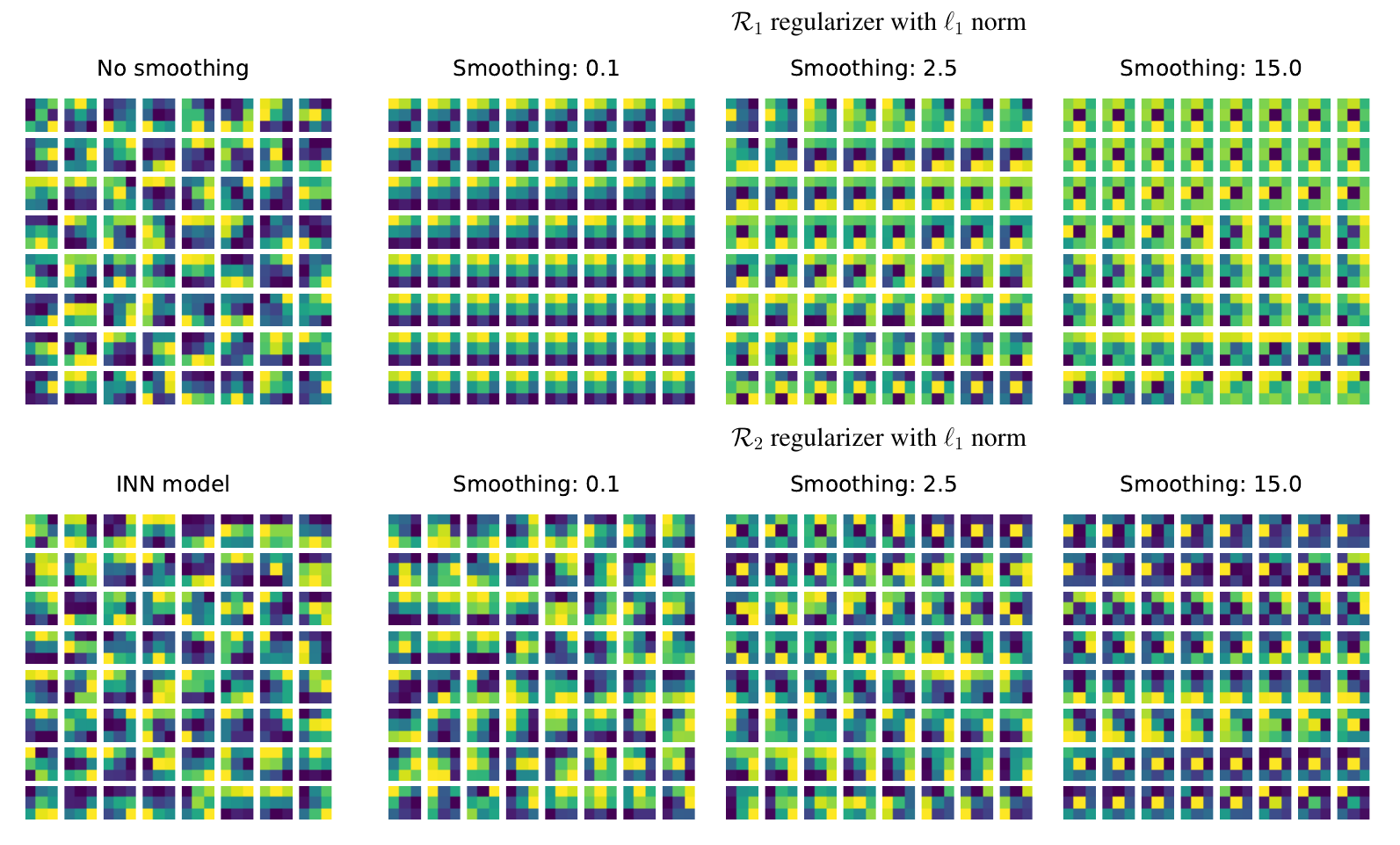}
    \caption{Inspecting the weighs of the first input channel of the ``layers.0.0.conv1'' layer of Resnet18 for different smoothing factors.
    The $64$ output kernels of the layer are sorted from left to right and top to bottom.
    For reference, the weights of the same layer for a non-smooth Resnet18 (top left) and an INN model (bottom left) are shown.
    We see that our regularization naturally imposes structures desired for the compression.
    }
    \label{fig:weights}
\end{figure}

Visual comparison of the weights of the first input channel of one of the ResNet layers (highlighted in Figure \ref{fig:weights}) confirms our findings. We compare the non-smooth ResNet, the INN model and the network trained with $\mathcal{R}_1$ regularizer (top row) and $\mathcal{R}_2$ regularizer (bottom row) for three different smoothing factors. The 64 output channels are ordered from left to right and top to bottom. We observe the expected behavior. For the network trained without smoothing, the channels weights are non-smooth. The INN model enforces more smoothness while being less smooth than both the weights trained with $\mathcal{R}_1$ and $\mathcal{R}_2$ regularization term. The networks trained with regularization terms yield the expected behavior, i.e., the $\mathcal{R}_1$ regularizer penalizes sharp changes and sudden jumps leading to smooth and slow changing values whereas the $\mathcal{R}_2$ regularizer penalizes curvature and oscillations but encourages linear functions. For both regularizers, smoothness increases with increasing smoothness factors. 

\begin{figure}
    \centering
    \includegraphics[width=.65\linewidth]{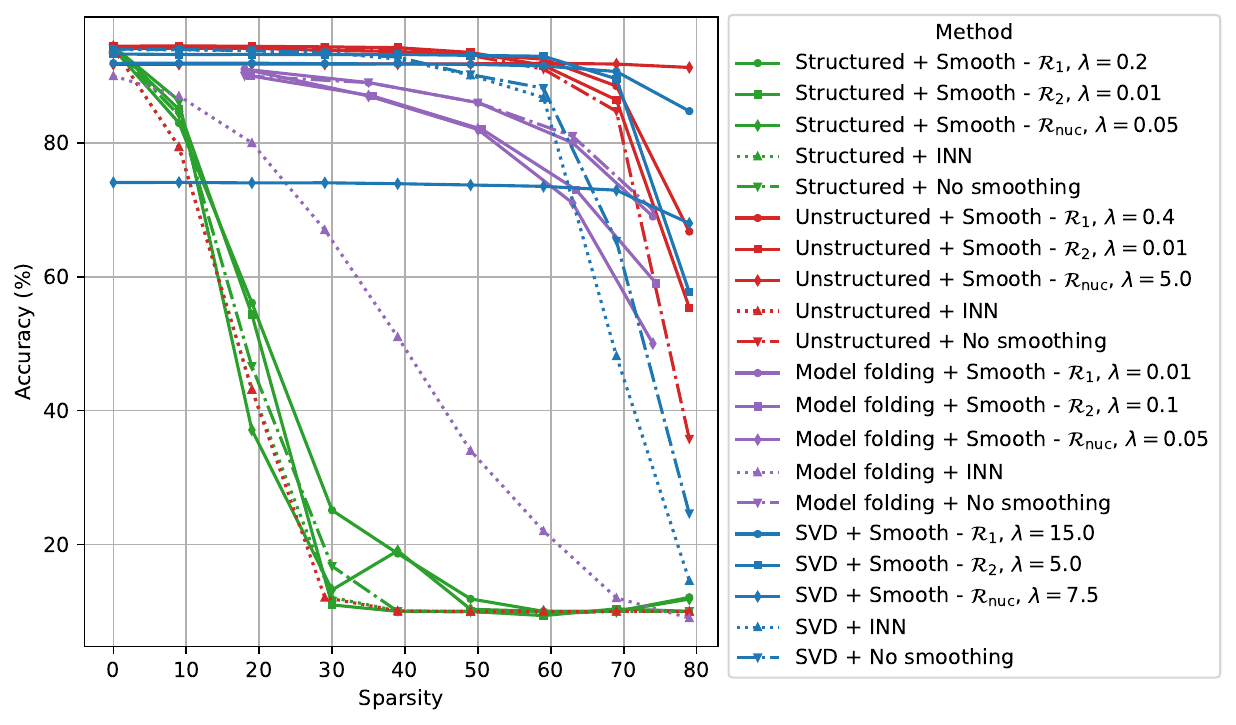}
    \caption{Benchmarking results of best performing smooth Resnet18 against INN and non-smooth Resnet18 models. We test SVD compression, structured $L_1$ pruning, unstructured $L_1$ pruning and model folding \cite{wangforget}. For sparsity levels $\geq 70\%$ and for a smoothing factor of $15.0$, our combination of smoothness and SVD-compression outperforms all considered competitors, with the $\mathcal{R}_1$ regularizer reaching up to $91\%$ and $85\%$ accuracy at sparsity levels $70\%$ and $80\%$, respectively.
    }
    \label{fig:cifar_all}
\end{figure}

We now apply the SVD compression pruning method to the networks trained with smooth weight learning. 
Figure \ref{fig:cifar_all} highlights the results for the best performing ResNet18 architecture trained with an $\mathcal{R}_1$ regularization term (solid line with circles), an $\mathcal{R}_2$ regularization term (solid line with squares) and, for the sake of completeness, the $\mathcal{R}_\text{nuc}$ regularizer (solid line with diamonds) compared to an INN model \cite{solodskikh2023integral} (dashed line with triangles) and a non-smooth version of the network trained without regularization term (long dashed line with triangles). 
As ``best performing'', we describe the method with the highest accuracy at sparsity $80\%$.
At a first glance, $\mathcal{R}_\text{nuc}$ with unstructured pruning and a large smoothing factor $(\lambda = 5.0)$, appears to be the best method.
However, as discussed in Sec. \ref{sec:related_work}, unstructured pruning does not decrease the size of the weight tensors.
In addition, calculating the nuclear morn as regularization method is computationally intensive and increases the training time by a factor of more than $5$.
%Therefore we focus our following discussion on structured pruning methods, and in particular our SVD pruning with $\mathcal{R}_1$ and  $\mathcal{R}_2$ regularizers.
A more detailed overview plot for all smoothing factors tested can be found in the appendix. We compare our SVD compression method (blue) against traditional pruning methods like structured $L_1$ pruning (green), unstructured $L_1$ pruning (red) and model folding \cite{wangforget} (purple). Structured $L_1$ pruning computes the $L_1$ norm of each output channel of a weight tensor, i.e., the dimension we would like to prune in, arranges the channels by their norm in descending order and removes as many channels as required to reach the desired sparsity level. Unstructured $L_1$ pruning however computes the magnitudes of each parameter individually, sorts them in descending order and sets the bottom $p$ percent to zero. 
While we found regularizing with a nuclear norm to be suitable for image compression using SVD, we surprisingly found it less effective for classification. 

The networks trained with smooth weight learning outperform INN as well as the networks trained without regularization term for any of the pruning methods for large sparsity levels. As indicated by the distribution of singular values, with increasing smoothing factor, the loss in performance when increasing the level of sparsity becomes smaller. We observe a similar behavior for $\mathcal{R}_1$ and $\mathcal{R}_2$ regularization terms with $\mathcal{R}_1$ outperforming its second-order counterpart for higher sparsity levels. Training the network with a smoothing factor of $15$ yields the best results -- achieving more than $90\%$ accuracy for a sparsity level of nearly $70\%$. For $80\%$ sparsity, the network trained with an $\mathcal{R}_1$ regularization term still reaches nearly $85\%$ accuracy on the CIFAR10 test dataset.

The combination of smooth weight learning and SVD compression outperforms every pruning method tested for both its unregularized counterpart and an INN with the same underlying network architecture. Smooth weight learning and SVD compression are especially powerful for large sparsity levels of more than $70\%$.

%% file: sec/5_conclusion.tex
\section{Conclusion}
\label{sec:conclusion}
In this paper, we introduce smooth weight learning in combination with singular value decomposition compression. Smooth weight learning imposes smoothness of the weights of the network by adding a regularization term to the loss function during training. Exploiting the smoothness of the weights, we propose singular value decomposition compression -- a pruning method that replaces the network's weight matrices by low-rank approximations. We demonstrate that the combination of smooth weight learning and SVD compression enables state-of-the-art compression on an implicit neural representation and image classification task -- without requiring any fine-tuning. We show that our method is especially powerful for high sparsity levels, reaching up to $91\%$ accuracy on a smooth ResNet18 for CIFAR10 classification with $70\%$ fewer parameters.
\paragraph{Limitations.} While our regularizers make fine-tuning-free compression via the SVD possible, the factorization of $W$ into two matrices may require a significant pruning of the rank before the number of parameters reduces, particularly for (near) quadratic $W$. Moreover, regularizers always introduce a hyperparameter for balancing regularity with the problem-specific loss. 

\section*{Acknowledgements}
C.R. acknowledges support from the Cantab Capital Institute for the Mathematics of Information (CCIMI) and the EPSRC grant EP/W524141/1. N.K.M. acknowledges support of the Deutsche Forschungsgemeinschaft (DFG, German Research Foundation), project number 534951134. J.L. and M.M. acknowledge support by the German Research Foundation research unit 5336 Learning to Sense and by the Lamarr Institute for Machine Learning and Artificial Intelligence. A.B. acknowledges the the Accelerate Programme for Scientific Discovery. C.B.S. acknowledges support from the Philip Leverhulme Prize, the Royal Society Wolfson Fellowship, the EPSRC advanced career fellowship EP/V029428/1, EPSRC grants EP/S026045/1 and EP/T003553/1, EP/N014588/1, EP/T017961/1, the Wellcome Innovator Awards 215733/Z/19/Z and 221633/Z/20/Z, the European Union Horizon 2020 research and innovation programme under the Marie Skodowska-Curie grant agreement No. 777826 NoMADS, the Cantab Capital Institute for the Mathematics of Information and the Alan Turing Institute.

%% file: sec/6_appendix.tex
%%%%%%%%%%%%%%%%%%%%%%%%%%%%%%%%%%%%%%%%%%%%%%%%%%%%%%%%%%%%
\appendix
\renewcommand{\thetable}{\Roman{table}}
\renewcommand{\thefigure}{\Roman{figure}}

\section*{Supplementary Material}

We provide further technical details on the experiments reported in the main text. 
This includes a description of hyperparameters and preprocessing settings for training the models in \Cref{sec:supp_seetings}, as well as a detailed analysis of the reported results in \Cref{sec:supp_results}.

%%%%%%%%%%%%%%%%%%%%%%%%%%%%%%%%%%%%%%%%%%%%%%%%%%%%%%%%%%%%

\section{Experimental Details}
\label{sec:supp_seetings}
This section details the hyperparameters, settings and data preprocessing steps used for the experiments in \Cref{sec:experiments}. 
\subsection{Implicit neural representations}
Table \ref{tab:supp_overview_training_settings_image_compression} provides an overview of the hyperparameters and training setting for the implicit neural representation experiments. We trained the models on a Nvidia GeForce RTX 5090.
\begin{table*}[h!]
    \centering
    \caption{Overview of training settings for training the WIRE models for image compression.}
    \begin{tabular}{l r}
    \toprule
      \textbf{Parameter}   &  \textbf{Value} \\ \midrule
      Loss function & Mean Squared Error \\ 
      Optimizer & Adam \\ 
      Learning rate & 0.01\\ 
      Non-linearity & WIRE\\
      Input features & 2 \\
      Output features & 3\\
      Learning rate scheduler & Cosine annealing to reduce it to $0.1$ times initial learning rate\\
      Iterations & 2000 \\ 
      Hidden features & 512 \\
      Smoothness & $\{0.001, 0.005, 0.01, 0.05, 0.1, 0.5, 1.0, 5.0\}$\\
      \\\bottomrule
    \end{tabular}
    \label{tab:supp_overview_training_settings_image_compression}
\end{table*}

\subsection{Image classification}
An overview of experimental details and hyperparameters for the image classification experiments on the CIFAR10 dataset \cite{krizhevsky2009learning} can be found in Table \ref{tab:supp_overview_training_settings_image_classification}. We use the standard train/test split. Training the models on a Nvidia Quadro RTX 6000 GPU took 2.5 hours per experiment. 
\begin{table*}[h!]
    \centering
    \caption{Overview of training settings for training the ResNet18 models for image classification.}
    \begin{tabular}{l r}
    \toprule
      \textbf{Parameter}   &  \textbf{Value} \\ \midrule
      Loss function & Cross Entropy Loss \\ 
      Optimizer & SGD \\ 
      Learning rate & 0.01\\ 
      Momentum & 0.9\\
      Weight decay & $5 \cdot 10^{-4}$\\
      Non-linearity & ReLU\\
      Regularization & Batch norm \\
      Width & 64\\
      Depth & 18\\
      Channels & 3 \\
      Classes & 10\\
      Learning rate scheduler 1 & Warm start from $0.0$ to $0.1$ over the first 5 epochs \\ 
      Learning rate scheduler 2 & Cosine annealing to reduce it to $0$ over remaining 295 epochs\\
      Epochs & 300 \\ 
      Batch size & 128\\  
      Smoothness & $\{0.01, 0.05, 0.1, 0.2, 0.3, 0.4, 0.5, 1.0, 2.0, 2.5, 5.0, 7.5, 10.0, 15.0\}$\\
      Data augmentations & random crop $32 \times 32$ patches while zero padding with $4$ pixels\\
      \\\bottomrule
    \end{tabular}
    \label{tab:supp_overview_training_settings_image_classification}
\end{table*}

\begin{figure}[h]
    \centering
    \includegraphics[width=.9\linewidth]{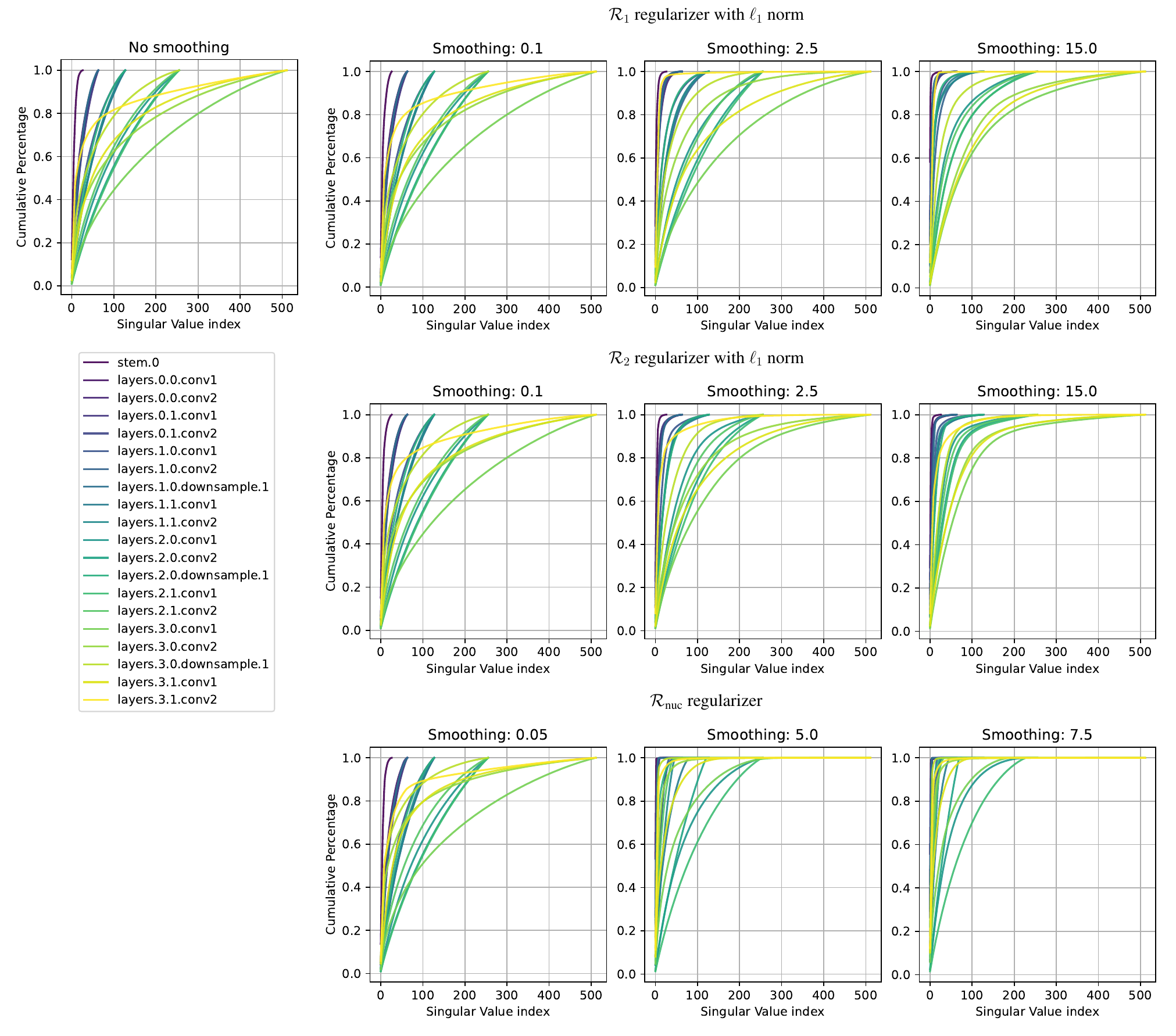}
    
    \caption{
    Completing Figure \ref{fig:svd_inspection} with the $\mathcal{R}_\text{nuc}$ regularizer.
    Compared to the $\mathcal{R}_1$ and $\mathcal{R}_2$ regularizers, the singular values of $\mathcal{R}_\text{nuc}$ cluster more sharply, which sharply reduces the networks' expressivity and my explain the accuracy drop fot large smoothness factors.
    }
    \label{fig:svd_inspection_nuclear}
\end{figure}

%%%%%%%%%%%%%%%%%%%%%%%%%%%%%%%%%%%%%%%%%%%%%%%%%%%%%%%%%%%%

\section{Further Numerical Results}
\label{sec:supp_results}
We complement the results of the image classification task with the ResNet model from \Cref{sec:results_image_classification}. Specifically, we provide an inspection of the singular values of models trained with the nuclear regularizer, as well as a visualization of the weights of one of their layers, completing \cref{fig:svd_inspection,fig:weights}.

The extended inspection of singular values is provided in Figure \ref{fig:svd_inspection_nuclear}. Compared to the $\mathcal{R}_1$ and $\mathcal{R}2$ regularizers, we observe that the $\mathcal{R}\text{nuc}$ regularizer clusters the singular values too sharply as the smoothing factor increases. This effect may explain the observed accuracy drop in smooth weight learning trained with the latter regularizer.

\begin{figure}[ht]
    \centering
    \includegraphics[width=1.\linewidth]{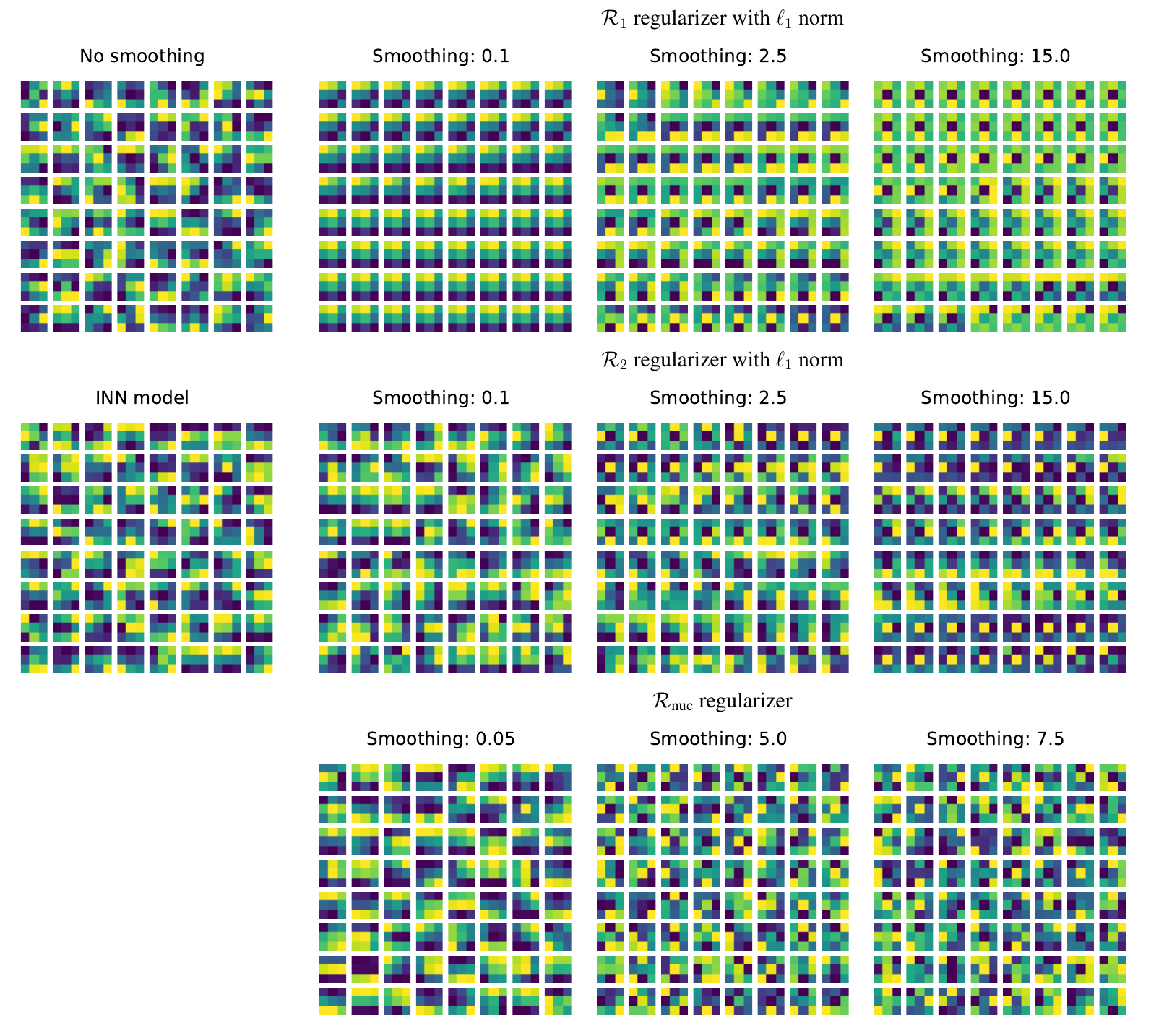}
    
    \caption{
    Completing Figure \ref{fig:weights} with the $\mathcal{R}_\text{nuc}$ regularizer.
    Compared to the $\mathcal{R}_1$ and $\mathcal{R}_2$ regularizers, with increasing smoothness factors the weights of $\mathcal{R}_\text{nuc}$ do not necessarily look smoother.
    }
    \label{fig:weights_nuclear}
\end{figure}

Examining the weight tensor of the “layers.0.0.conv1” layer in Figure \ref{fig:weights_nuclear}, we see that for a smaller smoothing factor ($\lambda = 0.05$), consecutive kernels appear to be rigid transformations of each other. However, for larger values of $\lambda$, no consistent trend can be observed.

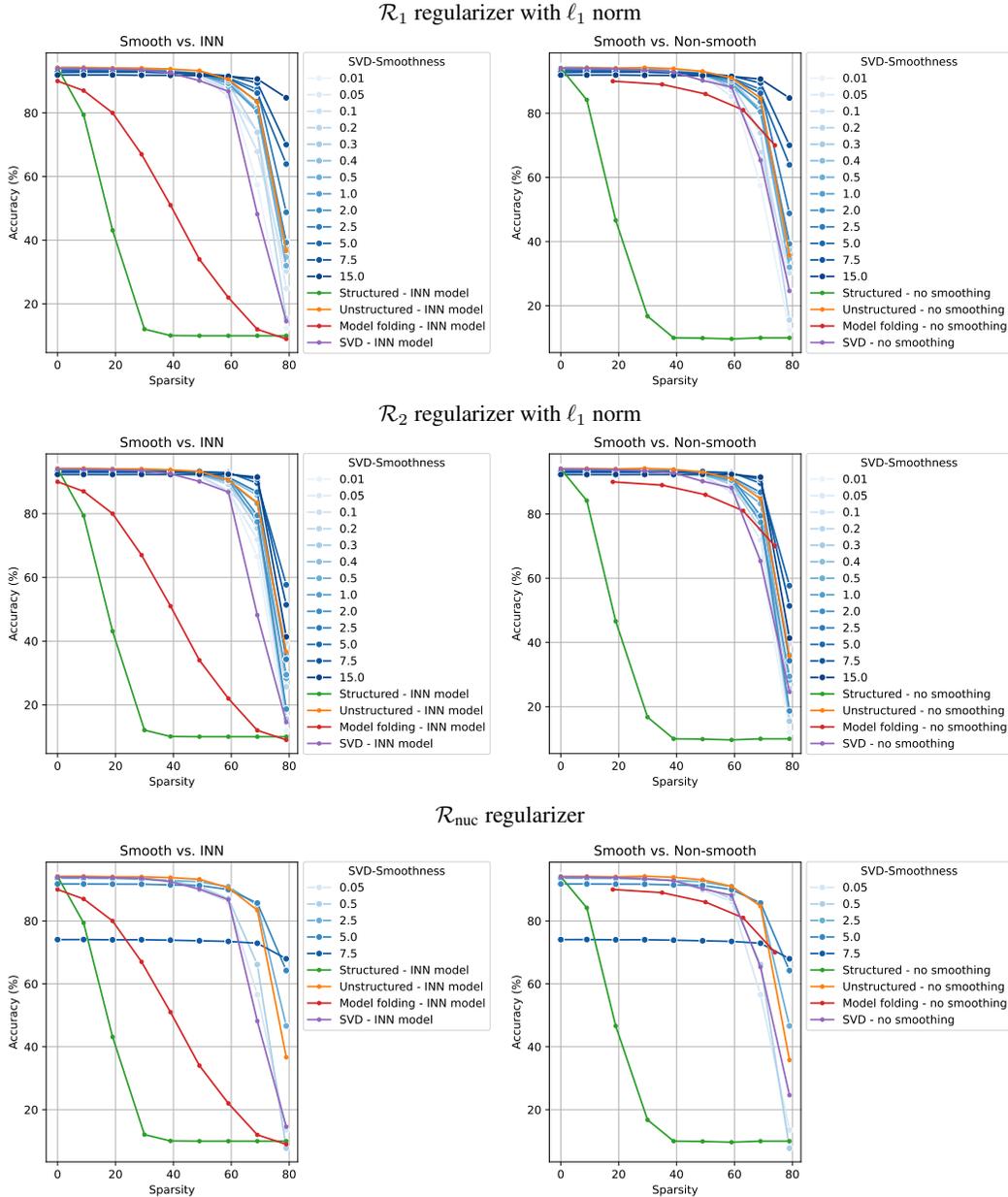
\begin{figure}[ht]
    \centering
    \begin{tikzpicture}
        \node[] at (0, -2.7) {
        \includegraphics[width=1.\linewidth]{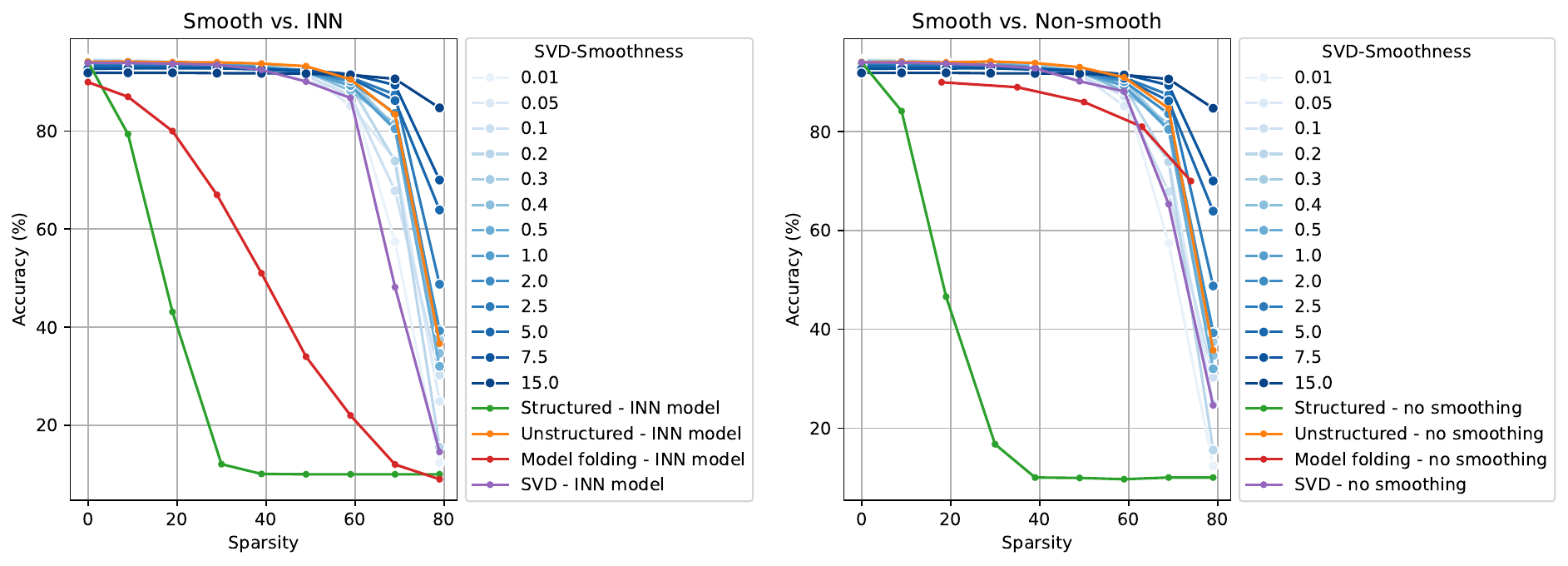}
        };
        \node[] at (0, -8.2) {
    \includegraphics[width=1.\linewidth]{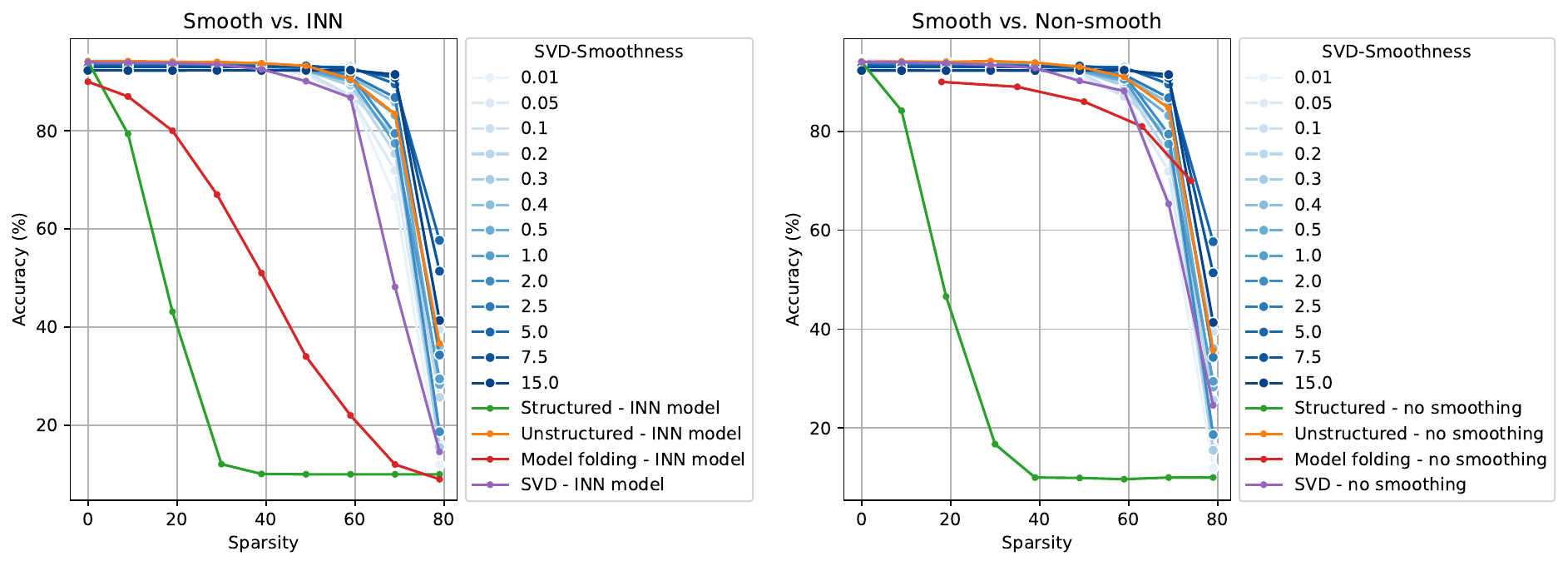}
        };
        \node[] at (0, -13.8) {
    \includegraphics[width=1.\linewidth]{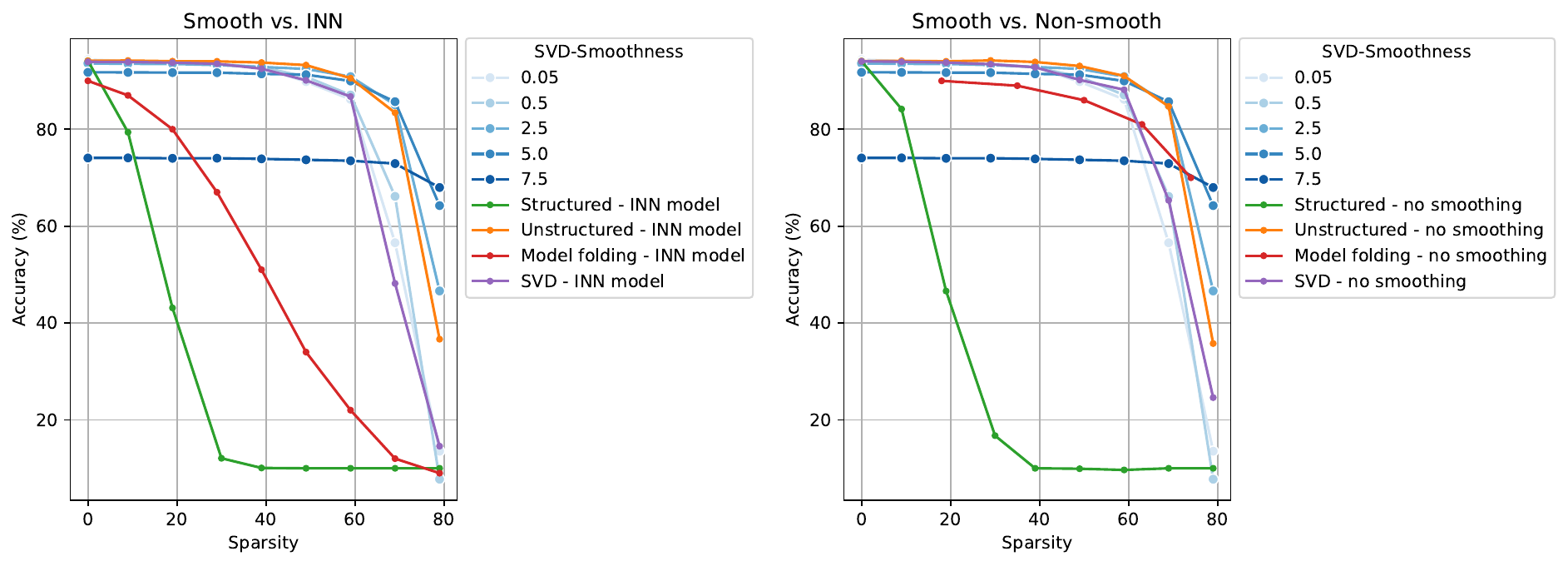}
        };
        \node[scale=.9] at (0, 0) {$\mathcal{R}_1$ regularizer with $\ell_1$ norm};
        \node[scale=.9] at (0, -5.5) {$\mathcal{R}_2$ regularizer with $\ell_1$ norm};
        \node[scale=.9] at (0, -11) {$\mathcal{R}_\text{nuc}$ regularizer};
    \end{tikzpicture}
    
    \caption{Benchmark results of the smooth Resnet18 and the SVD-compression against INN and non-smooth Resnet18 models, and three other compressions methods: Structure pruning, unstructured pruning and model folding \cite{wangforget}.
    \textbf{(Left)} Compression results on our smooth Resnet18 vs. INN models \cite{solodskikh2023integral}.
    \textbf{(Right)} Compression results on our smooth Resnet18 and non-smooth models.
    We see that at sparsity levels $\geq 70\%$ and for a smoothing factor of $15.0$, our combination of smoothness and SVD-compression outperforms all considered competitors, with the $\mathcal{R}_1$ regularizer reaching up to $91\%$ and $85\%$ accuracy at sparsity levels $70\%$ and $80\%$, respectively.
    The trend of reduced accuracy loss at higher sparsity levels with increasing smoothness factor can be verified also for the $\mathcal{R}_\text{nuc}$ regularizer.
    However, a large accuracy drop compared to other regularizes can be observed even without compression.
    }
    \label{fig:cifar_benchmark_extended}
\end{figure}

\paragraph{Model performance for different smoothing factors.}
\Cref{fig:cifar_benchmark_extended} compares the SVD-compression performances on our smooth models with different smoothing factor tested against INN models (left) and non-smooth models (right).
We additionally compare the combination of smooth weight learning and SVD compression to other pruning methods: Structure and un-structured $L_1$ pruning and model folding \cite{wangforget}. 

Clearly, we see that the accuracy of our compressed models improve with with higher smoothing factors, regardless on the chosen regularization term.
However, a too high smoothing factor reduces the expressivity of the models, making them less accurate than un-regularized models, even without compression.
This is particularly the case for $\mathcal R_\text{nuc}$ regularizer, which has an accuracy of $73\%$ without any compression.
This suggests the need to find optimal regularization factors to balance accuracy before and after compression.

INNs implicitly enforce smoothness along all continuous dimensions. 
We therefore compare our approach with an integral neural network for different pruning methods. 
Unstructured $L_1$ pruning achieves the best results and is on par with our method for sparsity levels up to $60\%$. 
The proposed combination of smooth weight learning and SVD compression, however, significantly outperforms INN for all four pruning methods (nearly $90\%$ vs. smaller than $85\%$ accuracy for our method and INN with the best performing pruning method for a sparsity level of $70\%$, respectively).

Comparing smooth weight learning to the same network architecture trained without regularization on the weights, for sparsity levels up to $60\%$, unstructured $L_1$ pruning and SVD compression perform on par with smooth weight learning pruned using SVD compression. 
For sparsity levels of more than $60\%$ however, the accuracy of the network without regularization drops off quickly. 
At a sparsity of $80\%$ of the original parameters, unstructured $L_1$ pruning achieves less than $40\%$ while SVD compression deteriorates the accuracy to less than $30\%$. 
Model folding achieves an accuracy of approximately $80\%$ for a sparsity level of around $60\%$, dropping off to $70\%$ accuracy for $75\%$ sparsity. 
The accuracy of the unregularized network using structured $L_1$ pruning resembles random guessing, i.e., $10\%$ for sparsity levels larger than $40\%$.

